\documentclass{article}

    \PassOptionsToPackage{numbers}{natbib}

     \usepackage[final]{paper}

\usepackage[utf8]{inputenc} 
\usepackage[T1]{fontenc}    
\usepackage{hyperref}       
\usepackage{url}            
\usepackage{booktabs}       
\usepackage{amsfonts}       
\usepackage{nicefrac}       
\usepackage{microtype}      

\usepackage{svg}
\usepackage{bm}

\usepackage[title]{appendix}
\usepackage{amsmath}
\usepackage{amssymb}
\usepackage{tabularx}

\usepackage{cleveref}

\title{The Search for Sparse, Robust Neural Networks}

\author{
    Justin Cosentino\thanks{Equal contribution.}~,~ Federico Zaiter\footnotemark[1]~,~Dan Pei\thanks{Corresponding authors.}~,~Jun Zhu\footnotemark[2]\\
    Department of Computer Science and Technology \\
    Tsinghua University\\
    Beijing, China \\
    \texttt{justin@cosentino.io},
    \texttt{zaitertf10@mails.tsinghua.edu.cn}, \\ \texttt{peidan@tsinghua.edu.cn}, \texttt{dcszj@mail.tsinghua.edu.cn}
}

\begin{document}

\maketitle

\begin{abstract}
    Recent work on deep neural network pruning has shown there exist sparse subnetworks that achieve equal or improved accuracy, training time, and loss using fewer network parameters when compared to their dense counterparts. Orthogonal to pruning literature, deep neural networks are known to be susceptible to adversarial examples, which may pose risks in security- or safety-critical applications. Intuition suggests that there is an inherent trade-off between sparsity and robustness such that these characteristics could not co-exist. We perform an extensive empirical evaluation and analysis testing the Lottery Ticket Hypothesis with adversarial training and show this approach enables us to find sparse, robust neural networks. Code for reproducing experiments is available here: \texttt{https://github.com/justincosentino/robust-sparse-networks}.
\end{abstract}

\section{Introduction}

It is well known that neural network pruning techniques can drastically reduce the number of parameters in trained networks without compromising accuracy, allowing for smaller and faster sparse representations of the original network \cite{lecun1990optimal}. Recent work suggests that dense, over-parameterized networks contain subnetworks whose architecture is responsible for the efficiency and accuracy of the original model \cite{frankle2018lottery, liu2018rethinking}. Retraining these substructures in isolation results in competitive performance and, at times, improved generalization.

 In recent years, the problem of adversarial examples has gained more and more attention \cite{szegedy2013intriguing, goodfellow2014explaining}. Preliminary empirical studies imply that there is a trade-off between network sparsity and adversarial robustness, but these studies do not attempt to take advantage of novel pruning techniques \cite{wang2018adversarial}. Aiming to determine the relationship between network architecture and adversarial robustness, we question whether over-parameterization is required in order to train robust networks, or if dense models also contain robust subnetworks responsible for the model's overall robustness. We conjectured that by using a pruning strategy similar to that proposed in the Lottery Ticket Hypothesis \cite{frankle2018lottery}, we could find subnetworks that may be responsible for the overall robustness of a model.

Orthogonal work has proven that there is an inherent tension between accuracy and robustness \cite{tsipras2018robustness}. The authors also found that the loss gradients in the input space align well with human perception. Not only are adversarially-trained gradients less noisy, but they appear to be more \textit{sparse}, suggesting that robust models may be well suited for retraining using sparse representations.

Deploying deep neural networks in embedded systems such as mobile phones, IoT devices, and smart wearable devices, which have limited storage and computing resources, has become more and more common \cite{Cheng2017ASO}. The use of overparametrized deep learning models in such scenarios poses several challenges when greater compression is required and performance must be maintained. At the same time, non-robust deep neural networks may deliver terrible results in critical applications. These scenarios would benefit greatly if sparse, robust neural networks can be successfully trained and deployed.

In these security-critical domains, the task of network verification is essential, i.e., the task of formally proving that no small perturbations of a given input can cause it to be misclassified by the network model. However, this process can be intractably slow even on small networks trained to classify tiny datasets such as MNIST. Recent work has shown that sparser networks drastically decrease the time taken to verify such networks, further motivating the need for robust, sparse representations \cite{xiao2018training}.

We perform an extensive empirical evaluation testing the Lottery Ticket Hypothesis (LTH) as part of the search for sparse, robust neural networks. We consider vision-centric classification tasks on the MNIST Digits and Fashion datasets. For the sake of comparison, we obtain sparse representations of the 300-100 Lenet architecture using different iterative pruning strategies. Adversarial training is performed with the FGSM \cite{goodfellow2014explaining} and PGD attacks \cite{madry2017towards}. We then empirically show that the Lottery Ticket Hypothesis can also find sparse, robust neural networks.

To summarize, our contributions are the following:
\begin{itemize}
    \item We build on the LTH by incorporating adversarial training into the pruning process.
    \item We show that this process can also find sparse, robust neural networks that train faster.
    \item We perform and present an extensive analysis of our experiments comparing the results from different levels of sparsity, pruning techniques, and adversarial attacks on the target datasets.
\end{itemize}

 The rest of this paper is organized as follows. We first introduce necessary background regarding network pruning and adversarial robustness in \Cref{section:prelim}. Secondly, we outline related work regarding the sparsity and robustness of neural networks in \Cref{section:related}. We then describe our experimental design in \Cref{section:expdesign} and present our results in \Cref{section:expresults}. Lastly, we present a discussion of the results, future work, and conclusion in \Cref{section:discussion}, \Cref{section:limitations_futurework}, and \Cref{section:conclusion} respectively.


\section{Preliminaries}
\label{section:prelim}

\subsection{Network Pruning}
\label{section:prelim-prune}

The pruning of neural networks can reduce the parameter count of trained networks by over 90\%, which decreases storage requirements and improves the performance of inference without compromising accuracy \cite{lecun1990optimal}. There are two types of strategies for pruning or making neural networks sparser: structured and unstructured. The structured approach takes into account the architecture of the model by defining the target, pruned architecture before pruning and then pruning at a layer or convolution channels level. This way, original structures such as convolution layers are preserved and it is not required to use specific libraries or hardware to realize pruning's benefits. The unstructured approach only looks at the weights level regardless of the architecture of the model. An example of an unstructured approach is training using $l_0$ regularization, which naturally yields a sparse model. In our experiments we use unstructured pruning strategies as described in \Cref{section:expdesign-prune}.

\subsection{The Lottery Ticket Hypothesis}
\label{section:prelim-lth}

The Lottery Ticket Hypothesis states that a randomly-initialized, dense neural network contains a subnetwork that is initialized such that---when trained in isolation---it can match the test accuracy of the original network after training for at most the same number of iterations \cite{frankle2018lottery}. The lottery ticket hypothesis predicts that there exists some mask $m$ where commensurate training time is less than the original network, commensurate accuracy is higher than the original network, and the masked network has fewer parameters than the original. Winning tickets can be identified by iteratively training a network and pruning its smallest-magnitude weights as described in the first strategy of \Cref{section:expdesign-prune}.

For consistency, we borrow notation from \cite{frankle2018lottery} for representing the sparsity of a mask $m$: $P_{m}=\frac{\|m\|_{0}}{|\theta|}$. For example, $P_{m}=25 \%$ when 75\% of the weights are pruned.

\subsection{Adversarial Robustness}
\label{section:prelim-robust}
Though deep neural networks provide state-of-the-art results for most machine learning tasks, recent work showed that these networks are vulnerable to adversarial examples: inputs that are nearly indistinguishable from natural data to the human eye and yet misclassified by the attacked neural network \cite{szegedy2013intriguing, papernot2016limitations}. The threat of adversarial attacks makes it difficult to deploy deep models in security- or safety-critical environments without some verifiable or provable level of robustness. A growing body of work \cite{goodfellow2014explaining, papernot2017practical, madry2017towards, carlini2017towards} shows the existence of white- and black-box attacks on neural networks, giving rise to the question: how can we learn models robust to adversarial inputs?

Recent work also showed that models with higher capacity, i.e., number of parameters, tend to be more robust to adversarial examples compared to lower capacity model of the same architecture \cite{kurakin2016adversarial}. The authors conjecture that increased model complexity helps build more robust models. Later work suggests a similar phenomena, finding that to reliably withstand strong adversarial attacks, networks require a significantly larger capacity than for only classifying benign examples \cite{madry2017towards}. They suggest that a robust decision boundary is significantly more complicated than its benign counterpart.

In order to test these hypotheses, we select two popular adversarial attacks: fast gradient sign method \cite{goodfellow2014explaining} and projected gradient descent \cite{madry2017towards}. We use each of these attacks to assess the robustness of the pruned models with and without adversarial training, which is known as an effective defense mechanism \cite{kurakin2016adversarial, madry2017towards}.

\subsubsection{Fast Gradient Sign Method}
\label{section:fgsm}
The Fast Gradient Sign Method (FGSM) \cite{goodfellow2014explaining} is a white-box attack method for generating adversarial examples. Given a natural example, it adds an imperceptibly small noise vector whose elements are equal to the sign of the elements of the gradient of the cost function with respect to the original input.

Let $\bm{\theta}$ denote the parameters of a neural network, $\bm{x}$ be the input to the network, $y$ be the labels associated with $\bm{x}$, and $J(\bm{\theta},\bm{x},y)$ be the cost used to train the network. FGSM linearizes the cost function around the current value of $\bm{\theta}$, obtaining an optimal constrained perturbation for the given input:
$$
\bm{\eta}=\epsilon \operatorname{sign}\left(\nabla_{\bm{x}} J(\bm{\theta}, \bm{x}, y)\right).
$$
Here, $\epsilon$ constrains the size of the perturbation: $\|\eta\|_{\infty}<\epsilon$. We then perturb the original image by adding this noise vector: $\bm{x}^{t+1} = \bm{x}^t + \bm{\eta}$.

\subsubsection{Projected Gradient Descent}
\label{section:pgd}
Projected Gradient Descent (PGD) \cite{madry2017towards} builds upon the aforementioned FGSM attack. Interpreting the FGSM attack as a simple one-step scheme, PGD represents a multi-step variant:
$$
\bm{x}^{t+1}=\Pi_{\bm{x}+\mathcal{S}}\left(\bm{x}^{t}+\alpha \operatorname{sign}\left(\nabla_{\bm{x}} J(\bm{\theta}, \bm{x}, y)\right)\right).
$$


\section{Related Work}
\label{section:related}
Recent work \cite{wang2018adversarial} also studies the robustness of pruned neural networks under adversarial attack, claiming that there exist trade-offs between standard classification accuracy, pruning rate, and adversarial robustness. The key difference between our work and \cite{wang2018adversarial} is that we evaluate the robustness of the pruning strategies used to find winning lottery tickets. We show that winning lottery tickets preserve not only a model's standard accuracy, but also adversarial robustness. We outline other distinctions here.

\paragraph{Model} \cite{wang2018adversarial} uses a 3-layer CNN, while we use a 300-100 Lenet architecture \cite{lecun1998gradient}, similarly to the fully-connected experiments in the Lottery Ticket Hypothesis \cite{frankle2018lottery}.

\paragraph{Pruning Strategy} \cite{wang2018adversarial} trains their model for 10 epochs, performs one-shot global weight pruning \cite{han2015learning, see2016compression} or one-shot global filter pruning \cite{li2016pruning}, and fine-tunes for another 10 epochs. We utilize the iterative pruning process described in the Lottery Ticket Hypothesis and outlined in \autoref{section:expdesign-prune}.

\paragraph{Adversarial Attacks} Similarly to \cite{wang2018adversarial}, we use the FGSM and PGD adversarial attacks to evaluate the robustness of the pruned network. \cite{wang2018adversarial}  also evaluates the model using a black-box attack \cite{papernot2017practical} and compares adversarially trained models on all attacks. We save this for future work.


\section{Experimental Design}
\label{section:expdesign}
In order to evaluate the robustness of Lottery Ticket networks, we create three separate pruning experiments, detailed in \autoref{section:expdesign-prune}. For each pruning strategy, we train the model with and without adversarial training using either the FGSM or PGD attack for 20 pruning iterations. A single pruning iteration consists of initializing the current iteration's parameters according to the pruning strategy, training for 50,000 iterations, and pruning some percent of the model to get an updated mask. We evaluate the model on both natural and adversarial examples from the entire validation and test set every 500 training iterations. Experimental results, unless otherwise noted, are averaged over five trials of each experiment. Any error metrics denote standard deviation.

We implemented the Lottery Ticket experiment using TensorFlow \cite{abadi2016tensorflow}, and we use the CleverHans \cite{papernot2018cleverhans} library to perform adversarial attacks on our TensorFlow Keras models. We trained and evaluated the models using NVIDIA GeForce GTX TITAN X GPUs.

\subsection{Iterative Pruning Strategies}
\label{section:expdesign-prune}
We examined three separate iterative pruning strategies in our experiments. ``Iterative pruning with resetting'' refers the core pruning strategy from the Lottery Ticket Hypothesis. We use the ``Iterative pruning with random resetting'' and ``Iterative pruning with continued training'' strategies as baselines to show the importance of the initial weights in the pruning process. Let $\mathcal{D}_{\theta}$ denote the distribution of initial parameters.

\paragraph{Strategy 1: Iterative pruning with resetting}
In this strategy, we reset the network to its original parameters $\theta_0  \sim \mathcal{D}_{\theta}$ after each training and pruning cycle.
\begin{enumerate}
    \item Randomly initialize a neural network $f(x ; m \odot \theta)$ where $\theta=\theta_{0} \sim \mathcal{D}_{\theta}$ and $m=1^{|\theta|}$ is a mask.
    \item Train the network for $j$ iterations, reaching parameters $m \odot \theta_{j}$.
    \item Prune $s\%$ of the parameters, creating an updated mask $m^\prime$ where $P_{m^{\prime}}=\left(P_{m}-s\right)\%$.
        \item Reset the weights of the remaining portion of the network to their values in $\theta_{0}$, i.e., let $\theta=\theta_{0}$.
    \item Let $m=m^{\prime}$ and repeat steps 2 through 4 until a sufficiently pruned network has been obtained.
\end{enumerate}
This pruning strategy is denoted ``original'' in the experimental results.

\paragraph{Strategy 2: Iterative pruning with random reinitialization}
In this strategy, we reinitialize the network to random parameters $\theta_{0}^{\prime} \sim \mathcal{D}_{\theta}$ after each training and pruning cycle.
\begin{enumerate}
    \item Randomly initialize a neural network $f(x ; m \odot \theta)$ where $\theta=\theta_{0}$ and $m=1^{|\theta|}$ is a mask.
    \item Train the network for $j$ iterations.
    \item Prune $s\%$ of the parameters, creating an updated mask $m^\prime$ where $P_{m^{\prime}}=\left(P_{m}-s\right)\%$.
    \item Reinitialize the weights of the remaining portion of the network to new random values $\theta_{0}^{\prime} \sim \mathcal{D}_{\theta}$, i.e., let $\theta=\theta_{0}^{\prime}$.
    \item Let $m=m^{\prime}$ and repeat steps 2 through 4 until a sufficiently pruned network has been obtained.
\end{enumerate}
This pruning strategy is denoted ``random'' in the experimental results.

\begin{table*}[!t]
    \centering
    \resizebox{\textwidth}{!}{
\begin{tabular}{ccccc}
\toprule
Sparsity & \multicolumn{2}{c}{FGSM} & \multicolumn{2}{c}{PGD} \\
        \cmidrule{2-5}
         &                                Natural &                                 Attack &                                Natural &                                 Attack \\
\midrule
   100.0 &  98.33 $\pm$ 00.06 / 98.25 $\pm$ 00.11 &  08.79 $\pm$ 00.40 / 94.30 $\pm$ 00.84 &  98.37 $\pm$ 00.03 / 98.38 $\pm$ 00.09 &  02.31 $\pm$ 00.88 / 45.32 $\pm$ 00.20 \\
    51.3 &  98.35 $\pm$ 00.07 / 98.13 $\pm$ 00.06 &  09.32 $\pm$ 00.22 / 97.09 $\pm$ 00.24 &  98.39 $\pm$ 00.04 / 98.07 $\pm$ 00.12 &  02.01 $\pm$ 01.02 / 60.14 $\pm$ 00.99 \\
    16.9 &  98.08 $\pm$ 00.06 / 97.90 $\pm$ 00.08 &  11.50 $\pm$ 01.38 / 97.46 $\pm$ 00.16 &  98.13 $\pm$ 00.02 / 97.73 $\pm$ 00.26 &  00.74 $\pm$ 00.11 / 59.91 $\pm$ 00.81 \\
    08.7 &  97.84 $\pm$ 00.10 / 97.54 $\pm$ 00.14 &  08.09 $\pm$ 01.80 / 97.02 $\pm$ 00.21 &  97.85 $\pm$ 00.08 / 97.20 $\pm$ 00.14 &  01.93 $\pm$ 00.71 / 57.60 $\pm$ 00.25 \\
    03.6 &  97.12 $\pm$ 00.09 / 97.00 $\pm$ 00.10 &  07.42 $\pm$ 02.97 / 95.23 $\pm$ 00.75 &  97.14 $\pm$ 00.07 / 95.58 $\pm$ 00.78 &  02.16 $\pm$ 01.21 / 48.81 $\pm$ 00.98 \\
    01.8 &  95.88 $\pm$ 00.35 / 95.49 $\pm$ 00.34 &  04.17 $\pm$ 01.29 / 90.15 $\pm$ 01.45 &  95.47 $\pm$ 00.23 / 92.67 $\pm$ 01.62 &  04.16 $\pm$ 01.28 / 38.23 $\pm$ 02.33 \\
\bottomrule
\end{tabular}
    }
    \vspace{.4em}
    \caption{MNIST Digits test accuracy of (normally trained / adversarially trained) using the Lenet 300-100 model pruned with the iterative Lottery Ticket approach on natural images paired with FGSM and PGD adversarial attacks after 30,000 training iterations. Sparsity labels are $P_m$--the fraction of weights remaining in the network after pruning. Values are the average and standard deviation of five trials.}
	\label{table:acc-digits}
\end{table*}

 \begin{table*}[!t]
    \centering
    \resizebox{\textwidth}{!}{
    \begin{tabular}{ccccc}
\toprule
Sparsity & \multicolumn{2}{c}{FGSM} & \multicolumn{2}{c}{PGD} \\
        \cmidrule{2-5}
         &                                Natural &                                 Attack &                                Natural &                                 Attack \\
\midrule
   100.0 &  89.04 $\pm$ 00.33 / 89.00 $\pm$ 00.28 &  08.51 $\pm$ 01.24 / 86.96 $\pm$ 00.26 &  88.91 $\pm$ 00.04 / 88.83 $\pm$ 00.11 &  03.78 $\pm$ 00.40 / 24.67 $\pm$ 00.71 \\
    51.3 &  88.93 $\pm$ 00.27 / 88.64 $\pm$ 00.29 &  09.03 $\pm$ 00.67 / 86.79 $\pm$ 00.34 &  89.11 $\pm$ 00.05 / 89.10 $\pm$ 00.12 &  04.44 $\pm$ 00.31 / 28.66 $\pm$ 00.57 \\
    16.9 &  88.45 $\pm$ 00.11 / 88.12 $\pm$ 00.16 &  08.79 $\pm$ 01.11 / 86.53 $\pm$ 00.34 &  88.40 $\pm$ 00.14 / 87.87 $\pm$ 00.32 &  04.99 $\pm$ 00.21 / 27.62 $\pm$ 00.97 \\
    08.7 &  88.02 $\pm$ 00.23 / 87.72 $\pm$ 00.25 &  09.85 $\pm$ 01.30 / 85.30 $\pm$ 00.35 &  87.50 $\pm$ 00.29 / 86.38 $\pm$ 00.96 &  05.44 $\pm$ 00.38 / 25.75 $\pm$ 02.20 \\
    03.6 &  87.41 $\pm$ 00.37 / 87.13 $\pm$ 00.25 &  09.37 $\pm$ 00.65 / 82.30 $\pm$ 00.69 &  87.19 $\pm$ 00.48 / 83.44 $\pm$ 01.33 &  05.69 $\pm$ 00.99 / 23.83 $\pm$ 01.00 \\
    01.8 &  86.13 $\pm$ 00.22 / 85.70 $\pm$ 00.20 &  08.96 $\pm$ 00.62 / 78.07 $\pm$ 01.40 &  86.21 $\pm$ 00.28 / 81.66 $\pm$ 01.35 &  05.22 $\pm$ 01.24 / 23.46 $\pm$ 00.79 \\
\bottomrule
\end{tabular}
    }
    \vspace{.4em}
    \caption{MNIST Fashion test accuracy of (normally trained / adversarially trained) using the Lenet 300-100 model pruned with the iterative Lottery Ticket approach on natural images paired with FGSM and PGD adversarial attacks after 30,000 training iterations. Sparsity labels are $P_m$--the fraction of weights remaining in the network after pruning. Values are the average and standard deviation of five trials.}
	\label{table:acc-fashion}
\end{table*}

\paragraph{Strategy 3: Iterative pruning with continued training}
In this strategy, we never reset the network to random parameters, continuing to train the current parameter set after each training and pruning cycle.
\begin{enumerate}
    \item Randomly initialize a neural network $f(x ; m \odot \theta)$ where $\theta=\theta_{0}$ and $m=1^{|\theta|}$ is a mask.
    \item Train the network for $j$ iterations.
    \item Prune $s\%$ of the parameters, creating an updated mask $m^\prime$ where $P_{m^{\prime}}=\left(P_{m}-s\right)\%$.
    \item Let $m=m^{\prime}$ and repeat steps 2 and 3 until a sufficiently pruned network has been obtained.
\end{enumerate}
This pruning strategy is denoted ``continued'' in the experimental results.

\subsection{Hyperparameters}
\label{section:expdesign-hp}
We use the following hyperparameters across all experiments. We keep these values as close to the original Lottery Ticket Hypothesis paper as possible. We use a Lenet 300-100 MLP \cite{lecun1998gradient}, which consists of three fully connected layers of size 300, 100, and 10, resulting in 266,200 trainable parameters. The first two hidden layers use a ReLU activation function and the final output layer uses a softmax. We do not use biases in any layer. Models are trained using the Adam optimizer \cite{kingma2014adam} and a learning rate of 1.2e-3. All models train for 50,000 iterations per pruning iteration and use a batch size of 60. The first two layers have a pruning rate of 20\%, while the output layer has a pruning rate of 10\%.

Normal training consists of minimizing the categorical cross entropy loss function on natural examples. Adversarial training minimizes a combination of categorical cross entropy loss on natural examples and adversarial examples with a 50/50 split.

We craft FGSM attacks in a $l^{\infty}$ ball of $\epsilon=0.3$. We craft PGD attacks with a step size of 0.05 for 10 iterations in a $l^{\infty}$ ball of $\epsilon=0.3$. All attacks are clipped to be within $[0,1]$.

\subsection{Datasets}
\label{section:expdesign-dataset}
We perform our experimental analysis on two datasets: MNIST Digits \cite{lecun1998gradient} and MNIST Fashion \cite{xiao2017fashion}. Both datasets consist of 60,000 training images and 10,000 test images. We further split both training sets into a 50,000 image training set and 10,000 image validation set. Each example is a 28x28x1 grayscale image associated with a label from one of ten classes. Images are preprocessed by normalizing grayscale values to $[0,1]$.


\section{Experimental Results}
\label{section:expresults}

\begin{figure*}[!ht]
    \centering
    \includegraphics[width=\textwidth]{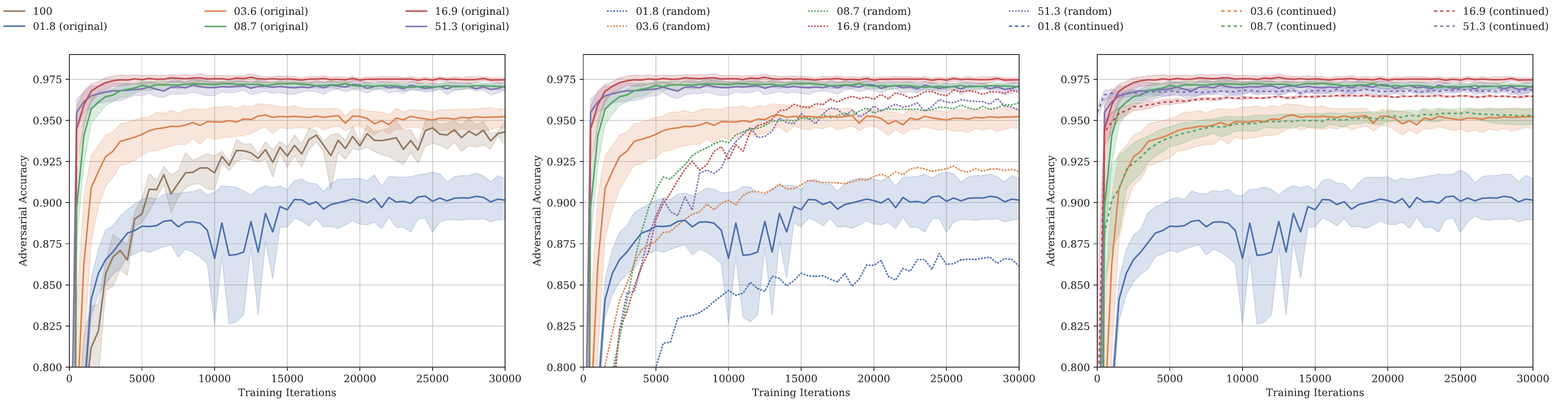}
    \caption{Lenet 300-100 adversarial test accuracy on MNIST Digits with the FGSM attack as training proceeds (left) and comparisons with the random (middle) and continued (right) pruning strategies. Labels are $P_m$--the fraction of weights remaining in the network after pruning. Each curve is the average of five trials and error bars are the standard deviation across trials.}
    \label{adv-acc-digits-fgsm}
\end{figure*}
\begin{figure*}[!ht]
    \centering
    \includegraphics[width=\textwidth]{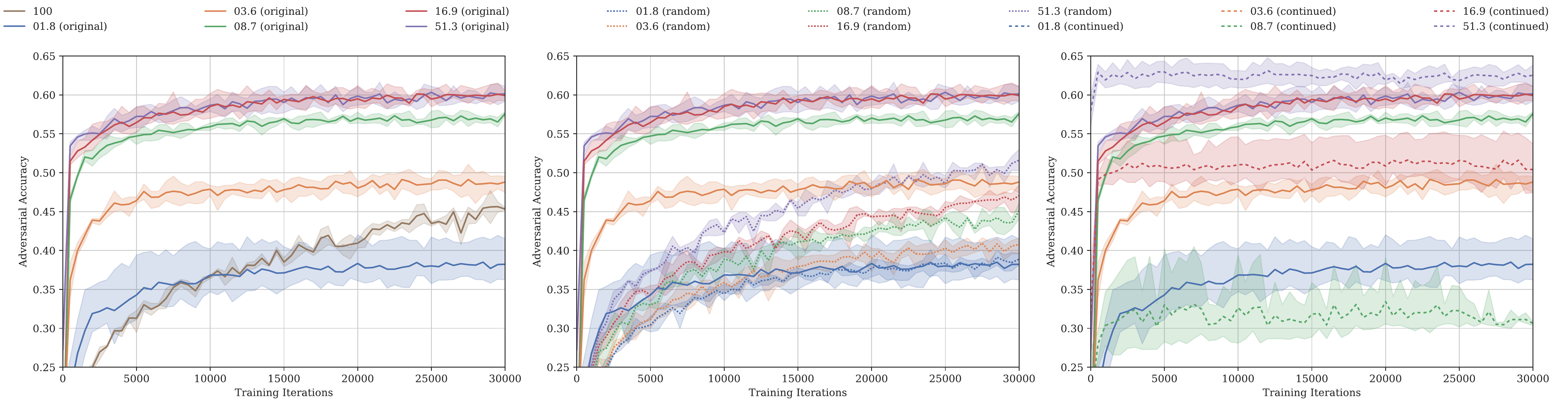}
    \caption{Lenet 300-100 adversarial test accuracy on MNIST Digits with the PGD attack as training proceeds (left) and comparisons with the random (middle) and continued (right) pruning strategies. Labels are $P_m$--the fraction of weights remaining in the network after pruning. Each curve is the average of five trials and error bars are the standard deviation across trials.}
    \label{adv-acc-digits-pgd}
\end{figure*}

\begin{figure*}[!ht]
    \centering
    \includegraphics[width=\textwidth]{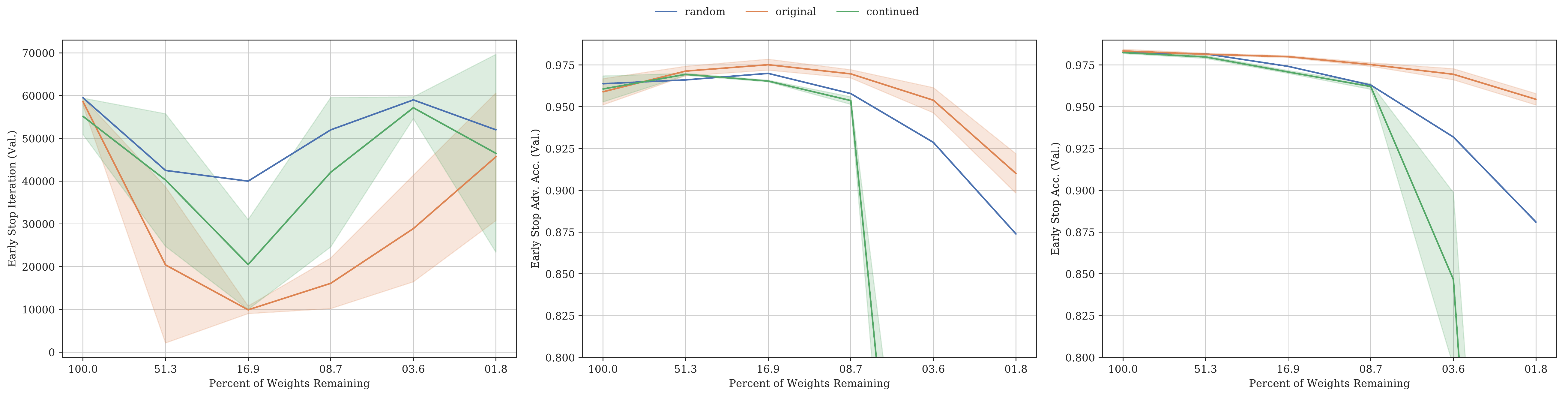}
    \caption{Early-stopping iteration (left), adversarial accuracy (middle), and natural accuracy (right) for each pruning strategy using Lenet 300-100 on MNIST Digits with the FGSM attack. Accuracy measures are taken at the early stopping iteration. Each curve is the average of five trials and error bars are the standard deviation across trials.}
    \label{early-stop-digits-fgsm}
\end{figure*}
\begin{figure*}[!ht]
    \centering
    \includegraphics[width=\textwidth]{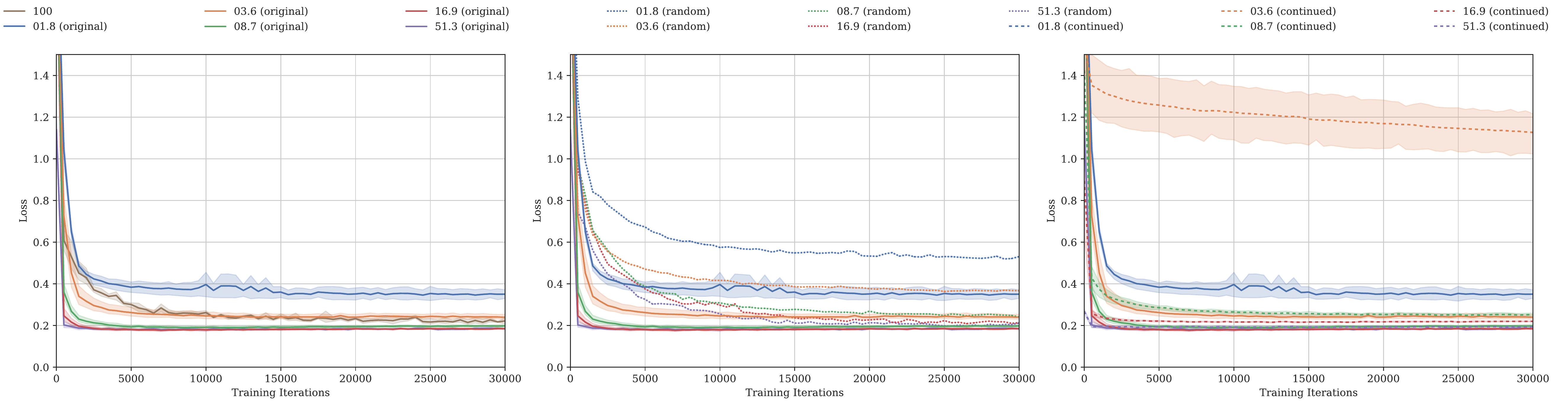}
    \caption{The adversarial validation loss data corresponding to \autoref{early-stop-digits-fgsm}, i.e., the adversarial validation loss for Lenet 300-100 on MNIST Digits with the FGSM attack as training proceeds (left) and comparisons with the random (middle) and continued (right) pruning strategies. Labels are $P_m$--the fraction of weights remaining in the network after pruning. Each curve is the average of five trials and error bars are the standard deviation across trials.}
    \label{test-loss-digits-fgsm}
\end{figure*}
\begin{figure*}[!ht]
    \centering
    \includegraphics[width=\textwidth]{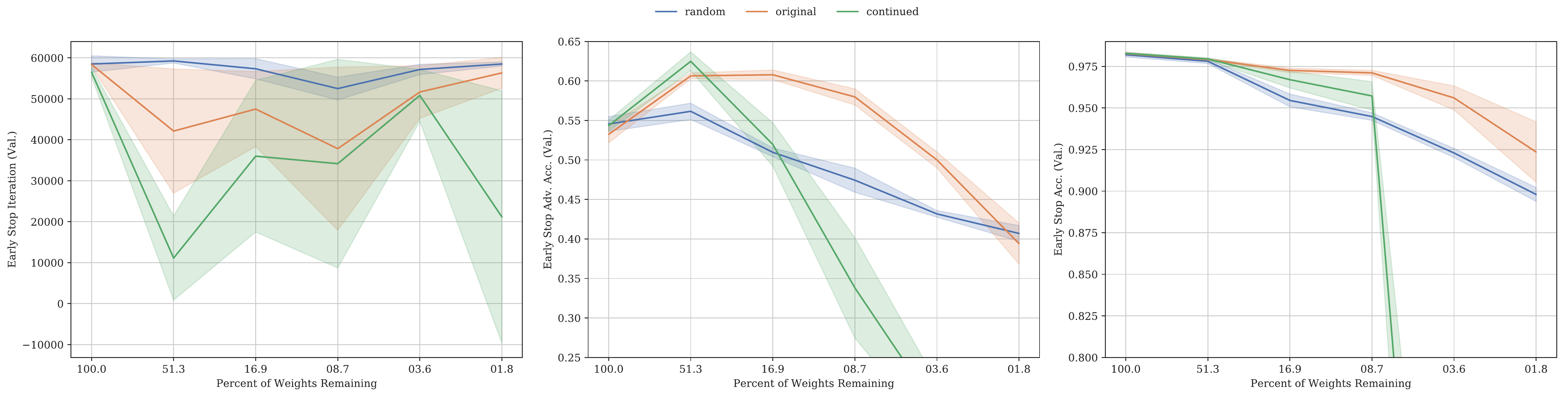}
    \caption{Early-stopping iteration (left), adversarial accuracy (middle), and natural accuracy (right) for each pruning strategy using Lenet 300-100 on MNIST Digits with the PGD attack. Accuracy measures are taken at the early stopping iteration. Each curve is the average of five trials and error bars are the standard deviation across trials.}
    \label{early-stop-digits-pgd}
\end{figure*}
\begin{figure*}[!ht]
    \centering
    \includegraphics[width=\textwidth]{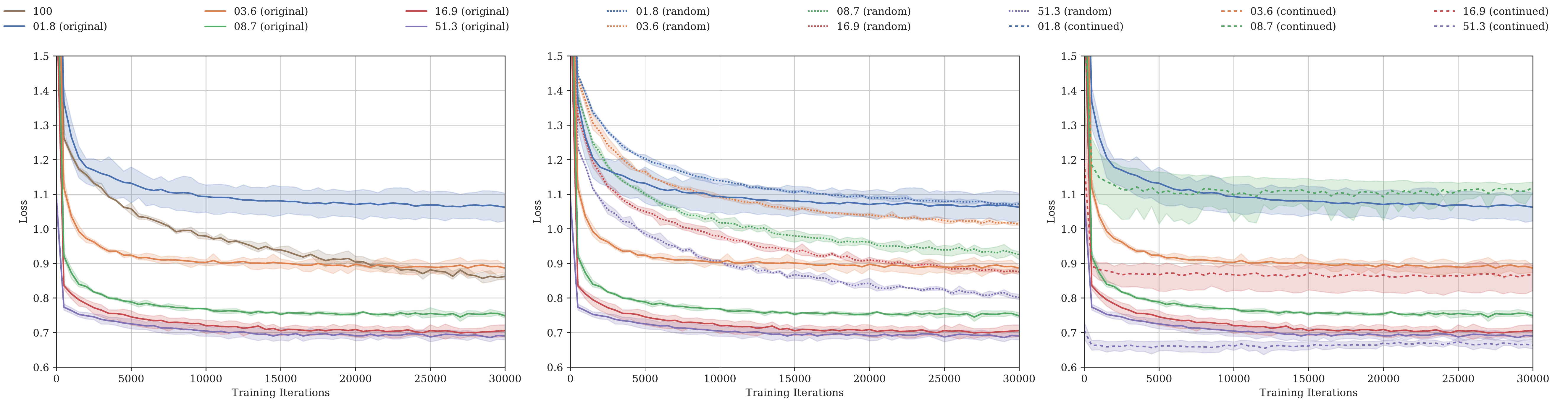}
    \caption{The adversarial validation loss data corresponding to \autoref{early-stop-fashion-pgd}, i.e., the adversarial validation loss for Lenet 300-100 on MNIST Fashion with the PGD attack as training proceeds (left) and comparisons with the random (middle) and continued (right) pruning strategies. Labels are $P_m$--the fraction of weights remaining in the network after pruning. Each curve is the average of five trials and error bars are the standard deviation across trials.}
    \label{test-loss-digits-pgd}
\end{figure*}

Both natural and adversarial test accuracy for the MNIST Digits and MNIST Fashion datasets are presented in \autoref{table:acc-digits} and     \autoref{table:acc-fashion}. In this section, we present figures for MNIST Digits results and include similar figures for MNIST Fashion results in Supplementary Section \ref{section:supp-mnist-fashion}.

\autoref{adv-acc-digits-fgsm} and \autoref{adv-acc-digits-pgd} show Lenet 300-100 adversarial test accuracy  as training proceeds on MNIST Digits with the FGSM attack and PGD attack, respectively. \autoref{early-stop-digits-fgsm} and \autoref{early-stop-digits-pgd} show the early-stopping iteration, early-stopping adversarial accuracy, and early-stopping natural accuracy for Lenet 300-100 on MNIST Digits with the FGSM attack and PGD attack, respectively. Early-stopping acts as a proxy measurement for the speed at which networks learn, denoting the training iteration at which the network would stop training given some early-stopping criterion. We use minimum validation loss as this metric in our experiments. We then report the adversarial and natural validation accuracy at this stopping point for various pruning levels. \autoref{test-loss-digits-fgsm} and \autoref{test-loss-digits-pgd} show the adversarial validation loss for Lenet 300-100 on the MNIST Digits with the FGSM attack and PGD attack, respectively.


\section{Discussion}
\label{section:discussion}

The Lottery Ticket Hypothesis pruning technique \cite{frankle2018lottery} has proven the existence of subnetworks that achieve better accuracy and generalization than the original dense network because of the fortuitous initialization of their weights. Using the same approach, we aim to test their hypothesis searching for subnetworks that can also achieve equal or better robustness at high levels of sparsity and compare it with different pruning techniques. This idea contradicts recent work that hypothesizes that models with higher capacity, i.e., number of parameters, tend to be more robust to adversarial examples compared to lower capacity model of the same architecture \cite{kurakin2016adversarial, madry2017towards} and suggests that fortuitous initialization and network architecture account for the overall robustness of the network. As shown in \autoref{adv-acc-digits-fgsm} and \autoref{adv-acc-digits-pgd}, we find robust lottery tickets that significantly outperform the original dense network on adversarial accuracy when pruned by up to 96.4\%. Additionally, as it is shown in \autoref{early-stop-digits-fgsm} and \autoref{early-stop-digits-pgd}, these sparser models also train faster than the original dense network achieving better results when the early stop best validation loss criterion is met. The best empirically found robust lottery ticket is at 16.9\% sparsity, significantly improving on the original dense network in adversarial accuracy while achieving similar results in natural accuracy.

Surprisingly, in specific scenarios, we found that continued training outperforms the Lottery Ticket Hypothesis in robustness. This is the case when using the PGD attack \cite{goodfellow2014explaining}, which is stronger than FGSM \cite{madry2017towards}. This can be appreciated in \autoref{adv-acc-digits-pgd} for the 51.3\% of sparsity in the continued training case. Likewise, when evaluating on the MNIST Fashion dataset using PGD attack, we see similar results having an even higher sparsity of 16.9\% as the most robust model. The common factor in both cases seems to be that the more complex the scenario, the better continued training will perform at higher levels of sparsity.

Even though the winning lottery tickets outperform all other approaches at any sparsity level on natural accuracy, the fact that continued training or other methods may outperform them on robustness in certain scenarios suggests it may be worth investigating its performance in order to find novel pruning techniques that could leverage the characteristics of both approaches and achieve even better results. In support of this, it is important to note that the random reinitialization pruning technique is outperformed by at least one of the other approaches in all cases.

Furthermore, winning lottery tickets constantly outperform all other approaches at the highest level of sparsity of 3.6\% and 1.8\%. In some cases even finding robust lottery tickets as in \autoref{adv-acc-fashion-fgsm}. This supports the hypothesis made by the authors of the Lottery Ticket paper \cite{frankle2018lottery} where up to a certain level of sparsity other pruning methods may still find subnetworks that can be retrained successfully, but beyond a certain point, this can only be achieved by methods that consider the fortuitous initialization of the weights in the pruning process.


\section{Limitations and Future Work}
\label{section:limitations_futurework}

We only consider vision-centric classification tasks on smaller datasets (MNIST Digits and MNIST Fashion) using a small fully-connected model (Lenet 300-100). We are constrained to these tasks as iterative pruning and adversarial training are computationally expensive, requiring twenty pruning iterations and numerous attack generations in order to evaluate the robustness of the model every 500 training iterations. Very recent work \cite{frankle2019lottery} has explored the lottery ticket hypothesis using larger models; we leave similar evaluations of larger models for robustness at scale for future work.

Though we drastically reduce parameter counts, the resulting unstructured, sparse architectures are not optimized for current libraries or hardware unlike their structured counterparts.

Lastly, our pruning strategies assume that small parameter weight corresponds to unimportant model parameters. This is a strong assumption.


\section{Conclusion}
\label{section:conclusion}

Contrary to empirical studies \cite{wang2018adversarial} and popular belief suggesting there is a trade-off between network sparsity and adversarial robustness, we find that overparameterization of the network is not required for network robustness. Winning lottery tickets not only account for the overall network's accuracy, but can also train faster and achieve similar---if not better---robustness with adversarial training. In specific scenarios, we found that continued training outperforms the Lottery Ticket Hypothesis in robustness, which suggests it may be worth investigating its performance in order to find novel pruning techniques that could leverage the characteristics of both approaches. In order to support these findings, we provide extensive empirical results comparing the Lottery Ticket Hypothesis approach with its alternatives.


{\small
\bibliographystyle{plain}
\bibliography{paper}
}


\clearpage

\title{Supplementary Material: The Search for Sparse, Robust Neural Networks}
\author{\vspace{-10ex}}
\maketitle

\renewcommand\appendixname{Supplement}
\renewcommand\appendixpagename{Supplement}
\begin{appendices}
\label{section:supp}

\section{Results for MNIST Fashion}
\label{section:supp-mnist-fashion}

\begin{figure*}[ht]
    \centering
    \includegraphics[width=\textwidth]{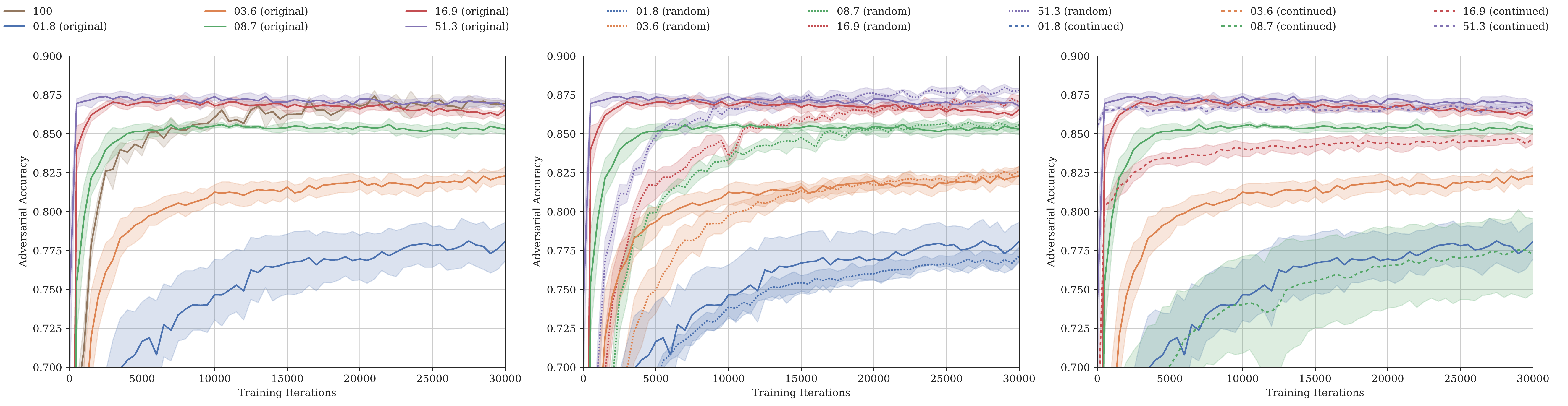}
    \caption{Lenet 300-100 adversarial test accuracy on MNIST Fashion with the FGSM attack as training proceeds (left) and comparisons with the random (middle) and continued (right) pruning strategies. Labels are $P_m$--the fraction of weights remaining in the network after pruning. Each curve is the average of five trials and error bars are the standard deviation across trials.}
    \label{adv-acc-fashion-fgsm}
\end{figure*}

\begin{figure*}[ht]
    \centering
    \includegraphics[width=\textwidth]{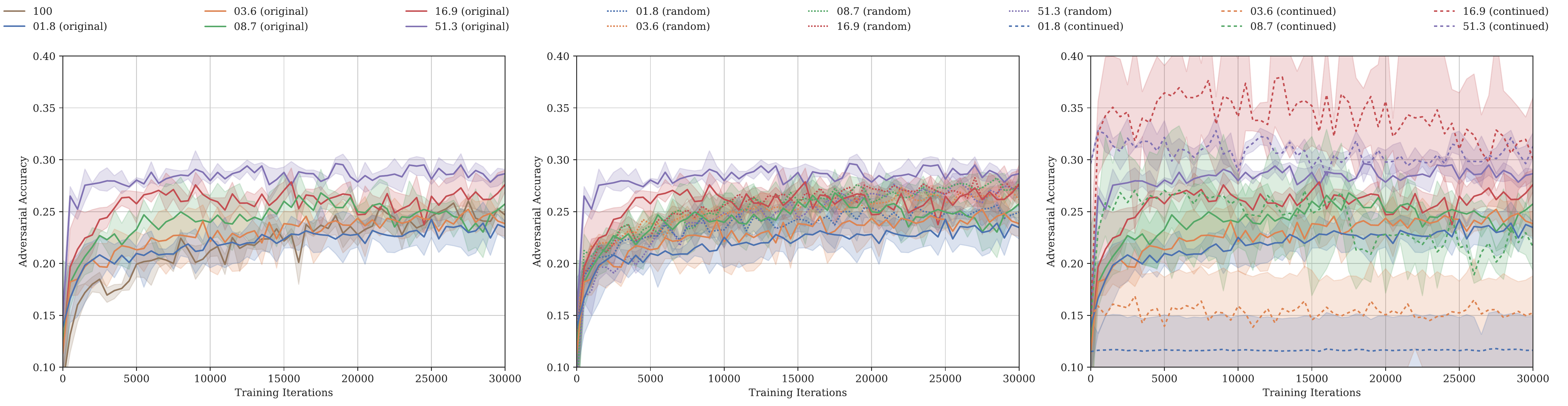}
    \caption{Lenet 300-100 adversarial test accuracy on MNIST Fashion with the PGD attack as training proceeds (left) and comparisons with the random (middle) and continued (right) pruning strategies. Labels are $P_m$--the fraction of weights remaining in the network after pruning. Each curve is the average of five trials and error bars are the standard deviation across trials.}
    \label{adv-acc-fashion-pgd}
\end{figure*}

\begin{figure*}[ht]
    \centering
    \includegraphics[width=\textwidth]{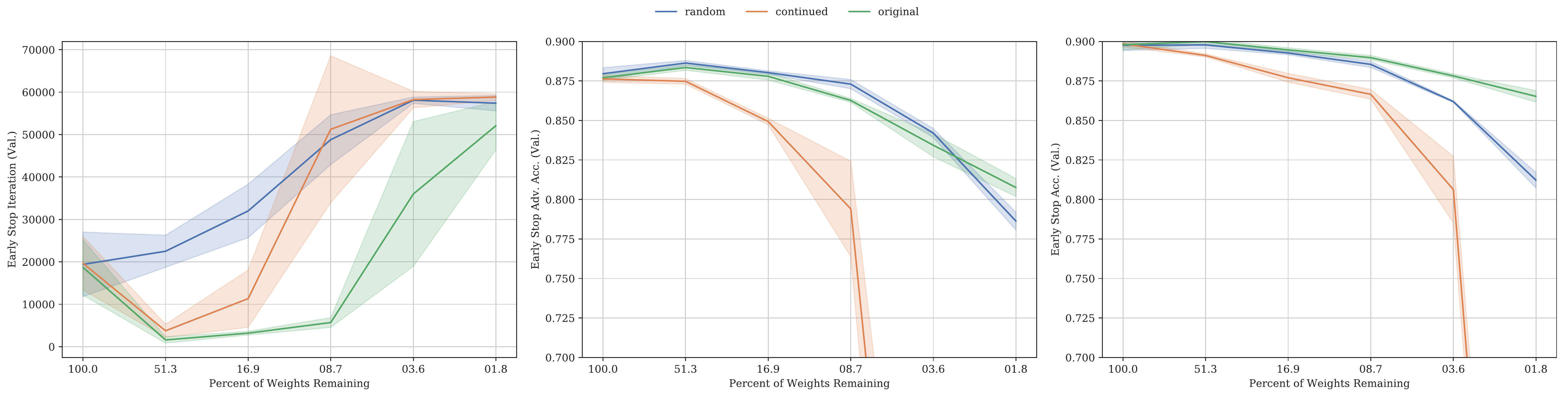}
    \caption{Early-stopping iteration (left), adversarial accuracy (middle), and natural accuracy (right) for each pruning strategy using Lenet 300-100 on MNIST Fashion with the FGSM attack. Accuracy measures are taken at the early stopping iteration. Each curve is the average of five trials and error bars are the standard deviation across trials.}
    \label{early-stop-fashion-fgsm}
\end{figure*}

\begin{figure*}[ht]
    \centering
    \includegraphics[width=\textwidth]{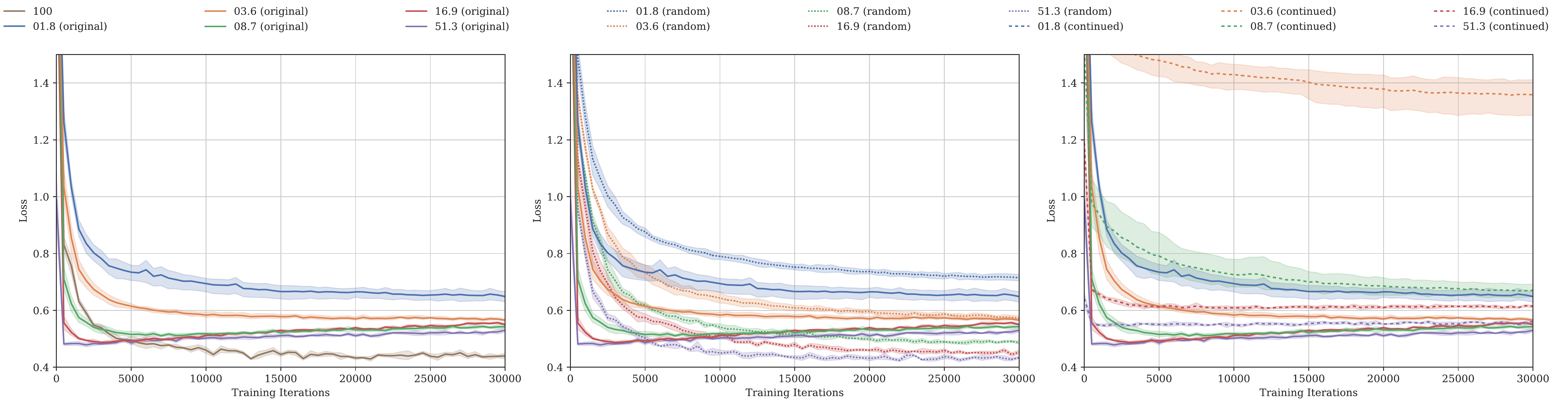}
    \caption{The adversarial validation loss data corresponding to \autoref{early-stop-fashion-fgsm}, i.e., the adversarial validation loss for Lenet 300-100 on MNIST Fashion with the FGSM attack as training proceeds (left) and comparisons with the random (middle) and continued (right) pruning strategies. Labels are $P_m$--the fraction of weights remaining in the network after pruning. Each curve is the average of five trials and error bars are the standard deviation across trials.}
    \label{test-loss-fashion-fgsm}
\end{figure*}

\begin{figure*}[ht]
    \centering
    \includegraphics[width=\textwidth]{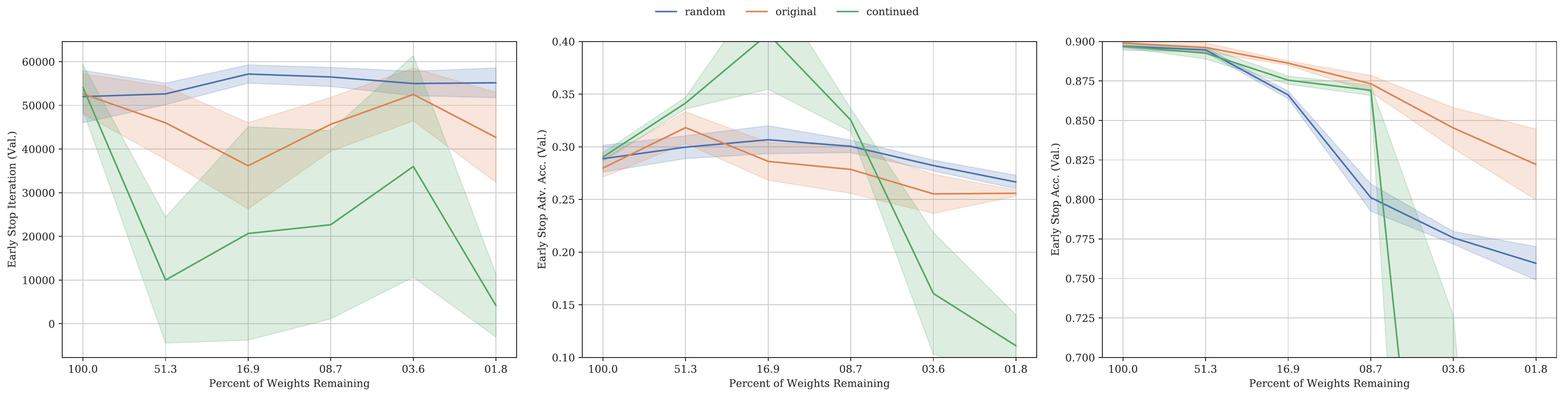}
    \caption{Early-stopping iteration (left), adversarial accuracy (middle), and natural accuracy (right) for each pruning strategy using Lenet 300-100 on MNIST Fashion with the PGD attack. Accuracy measures are taken at the early stopping iteration. Each curve is the average of five trials and error bars are the standard deviation across trials.}
    \label{early-stop-fashion-pgd}
\end{figure*}

\begin{figure*}[ht]
    \centering
    \includegraphics[width=\textwidth]{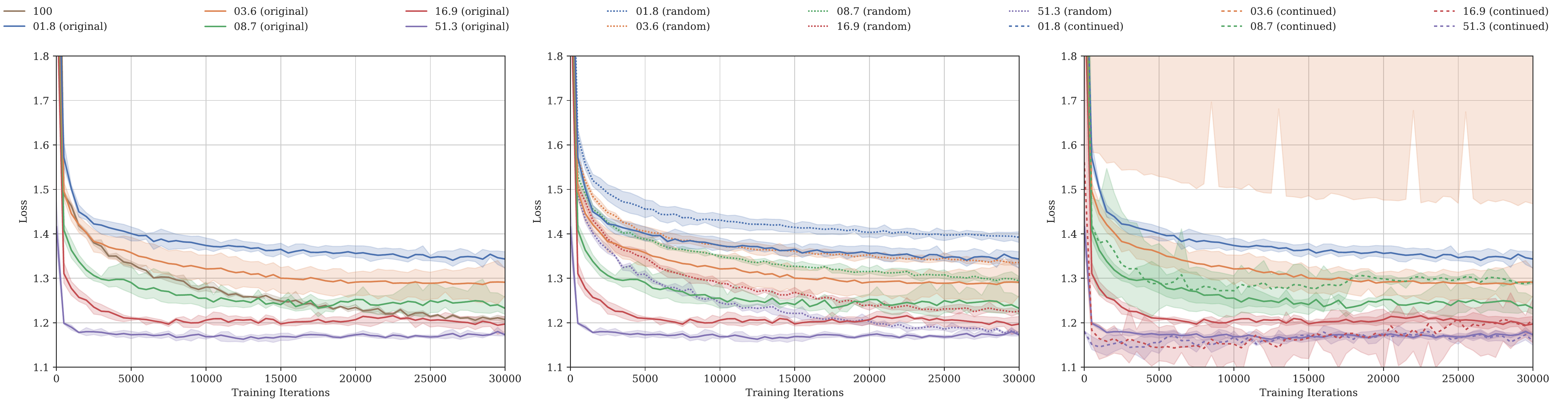}
    \caption{The adversarial validation loss data corresponding to \autoref{early-stop-fashion-pgd}, i.e., the adversarial validation loss for Lenet 300-100 on MNIST Fashion with the PGD attack as training proceeds (left) and comparisons with the random (middle) and continued (right) pruning strategies. Labels are $P_m$--the fraction of weights remaining in the network after pruning. Each curve is the average of five trials and error bars are the standard deviation across trials.}
    \label{test-loss-fashion-pgd}
\end{figure*}

In this section we present additional figures for the MNIST Fashion dataset. \autoref{adv-acc-fashion-fgsm} and \autoref{adv-acc-fashion-pgd} show Lenet 300-100 adversarial test accuracy  as training proceeds on MNIST Fashion with the FGSM attack and PGD attack, respectively. \autoref{early-stop-fashion-fgsm} and \autoref{early-stop-fashion-pgd} show the early-stopping iteration, early-stopping adversarial accuracy, and early-stopping natural accuracy for Lenet 300-100 on MNIST Fashion with the FGSM attack and PGD attack, respectively. Early-stopping acts as a proxy measurement for the speed at which networks learn, denoting the training iteration at which the network would stop training given some early-stopping criterion. We use minimum validation loss as this criterion in our experiments. We then report the adversarial and natural validation accuracy at this stopping point for various pruning levels. \autoref{test-loss-fashion-fgsm} and \autoref{test-loss-fashion-pgd} show the adversarial validation loss for Lenet 300-100 on the MNIST Fashion with the FGSM attack and PGD attack, respectively.


\section{Network Weight Distributions}
\begin{figure*}[ht]
    \centering
    \includegraphics[width=\textwidth]{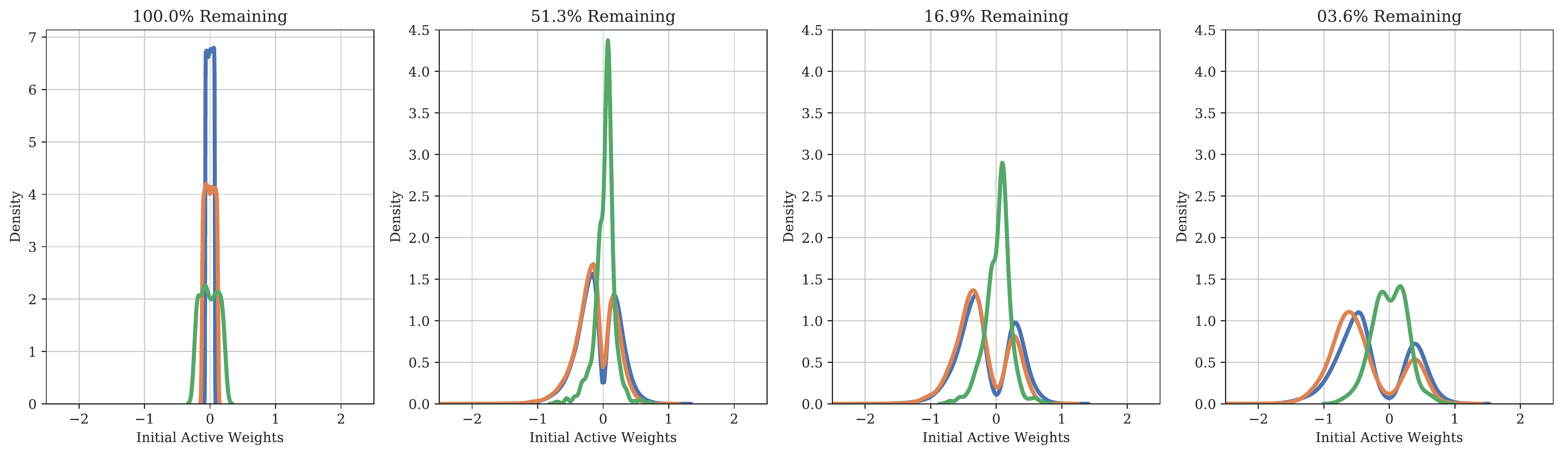}
    \caption{The distribution of initializations in winning tickets pruned to the levels specified in the titles of each plot using original reintialization. The blue, orange, and green lines show the distributions for the first hidden layer, second hidden layer, and output layer of the Lenet 300-100 for MNIST Digits with normal training. The distributions have been normalized so that the area under each curve is 1.}
    \label{fig:test-dist-digits-none}
\end{figure*}

\begin{figure*}[ht]
    \centering
    \includegraphics[width=\textwidth]{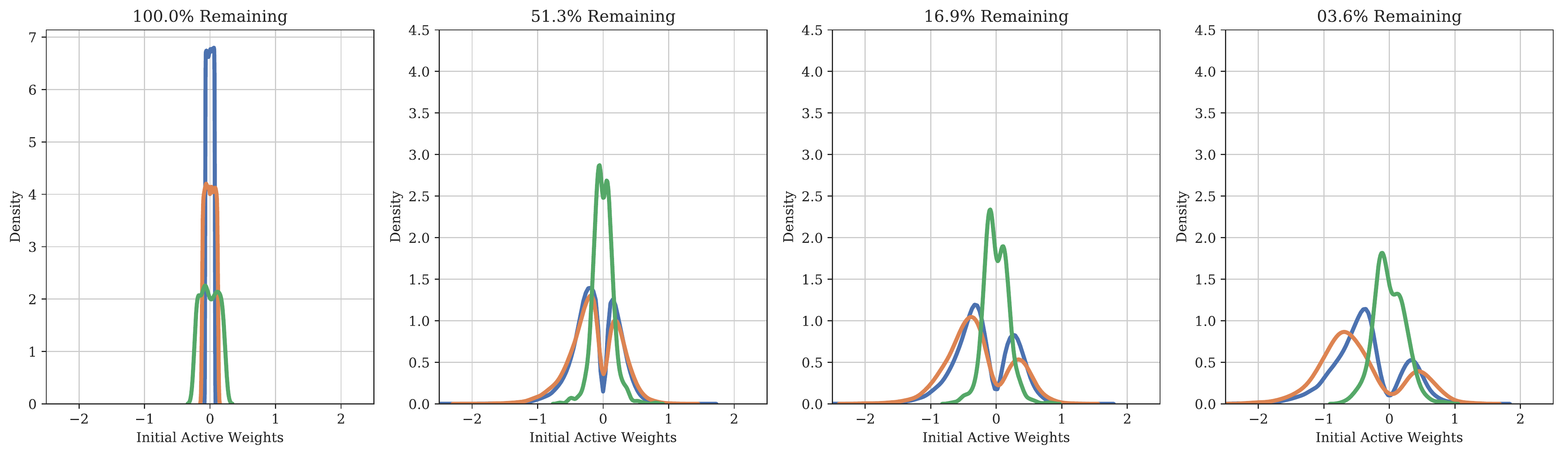}
    \caption{The distribution of initializations in winning tickets pruned to the levels specified in the titles of each plot using original reintialization. The blue, orange, and green lines show the distributions for the first hidden layer, second hidden layer, and output layer of the Lenet 300-100 for MNIST Digits with FGSM adversarial training. The distributions have been normalized so that the area under each curve is 1.}
    \label{fig:test-dist-digits-fgsm}
\end{figure*}

\begin{figure*}[ht]
    \centering
    \includegraphics[width=\textwidth]{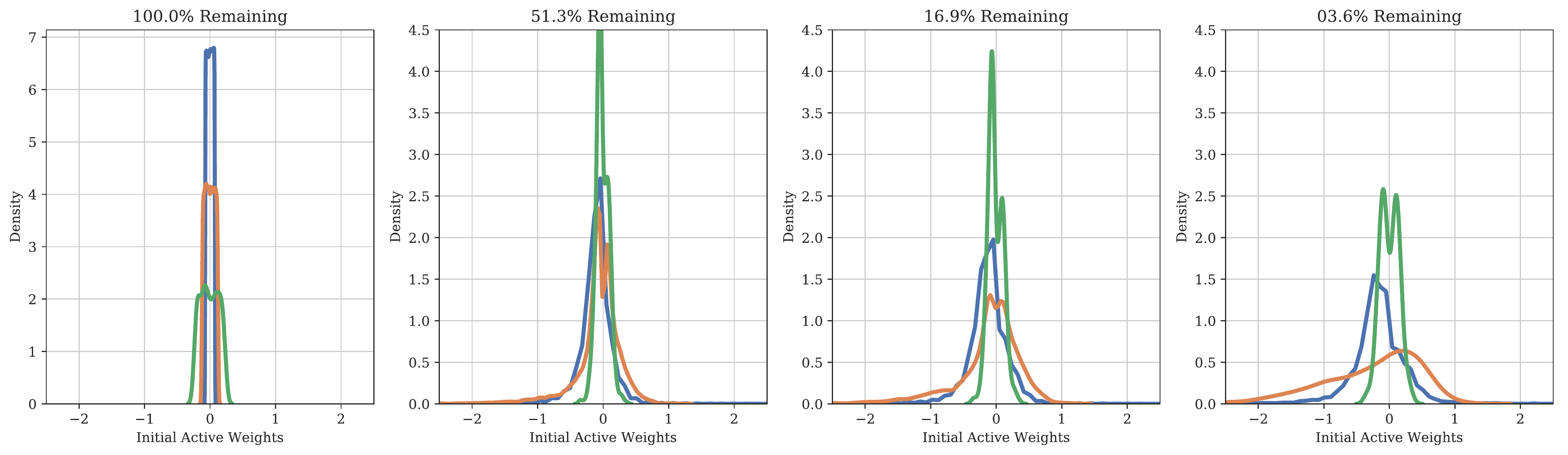}
    \caption{The distribution of initializations in winning tickets pruned to the levels specified in the titles of each plot using original reintialization. The blue, orange, and green lines show the distributions for the first hidden layer, second hidden layer, and output layer of Lenet 300-100 for MNIST Digits with PGD adversarial training. The distributions have been normalized so that the area under each curve is 1.}
    \label{fig:test-dist-digits-pgd}
\end{figure*}

\autoref{fig:test-dist-digits-none}, \autoref{fig:test-dist-digits-fgsm}, and \autoref{fig:test-dist-digits-pgd} show the distribution of initializations in winning tickets pruned to various sparsity levels using the original reinitialization strategy outlined in \autoref{section:prelim-prune} for natural training, FGSM adversarial training, and PGD adversarial training, respectively.


\end{appendices}

\end{document}